\documentclass[letterpaper, 10 pt, conference]{ieeeconf}  

\IEEEoverridecommandlockouts                              

\usepackage{graphics} 
\usepackage{epsfig} 
\usepackage{mathptmx} 
\usepackage{times} 
\usepackage{amsmath} 
\usepackage{amssymb}  
\usepackage{cite} 
\usepackage{mathtools} 
\usepackage{makecell}
\usepackage[hidelinks]{hyperref}
\usepackage[capitalise]{cleveref}

\crefname{equation}{}{}
\crefname{figure}{Fig.}{Figs.}

\usepackage{color} 
\usepackage{xcolor}
\definecolor{agreen}{rgb}{0.25, 0.51, 0.20}
\definecolor{better_blue}{rgb}{ 0.0627    0.5569    0.9098}

\DeclarePairedDelimiter{\norm}{\lVert}{\rVert}

\title{\LARGE \bf
Motion Planning for Agile Legged Locomotion using Failure Margin Constraints
}

\author{Kevin Green, John Warila, Ross L. Hatton, Jonathan Hurst
\thanks{This work was supported in part by NSF Grants 1314109-DGE, 1849343-IIS, 1653220-CMMI, and 1826446-CMMI \newline
\- All authors are affiliated with Collaborative Robotics and Intelligent Systems Institute, Oregon State University, Corvallis, OR, USA. Email
        {\tt\small greenkev@oregonstate.edu, johnwarila@gmail.com, \{ross.hatton, jonathan.hurst\}@oregonstate.edu }}%
}

\begin{document}

\maketitle
\thispagestyle{empty}
\pagestyle{empty}

\begin{abstract}
The complex dynamics of agile robotic legged locomotion requires motion planning to intelligently adjust footstep locations.
Often, bipedal footstep and motion planning use mathematically simple models such as the linear inverted pendulum, instead of dynamically-rich models that do not have closed-form solutions.
We propose a real-time optimization method to plan for dynamical models that do not have closed form solutions and experience irrecoverable failure.
Our method uses a data-driven approximation of the step-to-step dynamics and of a failure margin function.
This failure margin function is an oriented distance function in state-action space where it describes the signed distance to success or failure. 
The motion planning problem is formed as a nonlinear program with constraints that enforce the approximated forward dynamics and the validity of state-action pairs.
For illustration, this method is applied to create a planner for an actuated spring-loaded inverted pendulum model. 
In an ablation study, the failure margin constraints decreased the number of invalid solutions by between 24 and 47 percentage points across different objectives and horizon lengths.
While we demonstrate the method on a canonical model of locomotion, we also discuss how this can be applied to data-driven models and full-order robot models.

\end{abstract}

\section{Introduction}

The full-order dynamics of legged robots are too complex for real-time motion planning so motion planners often use simplified, reduced-order models of locomotion.
These models are able to reduce the state dimension while still capturing the core underactuated dynamics of locomotion.
Some of these models also have computationally efficient dynamics, particularly the closed-form dynamics of the linear inverted pendulum (LIP) model \cite{Feng2016LIPFootstep}.

Unfortunately, there is a trade-off to using some of these computationally efficient reduced-order models.
These models fail to exhibit features of locomotion which are known to relate to robust, efficient locomotion that are observed in humans and animals.
Energetically optimal gaits for bipedal robots include vertical center of mass oscillations and step frequency that varies with speed \cite{SmitAnseeuw2017RAMone}.
Neither of these features appear in LIP footstep planning.
The more complex spring loaded inverted pendulum (SLIP) model, not the LIP model, exhibits the center of mass dynamics and ground reaction forces of human walking and running \cite{Geyer2006}, as well as the disturbance response to drop steps in guineafowl \cite{Blum2014}.

\begin{figure} 
    \centering
    \includegraphics[width=0.80\columnwidth]{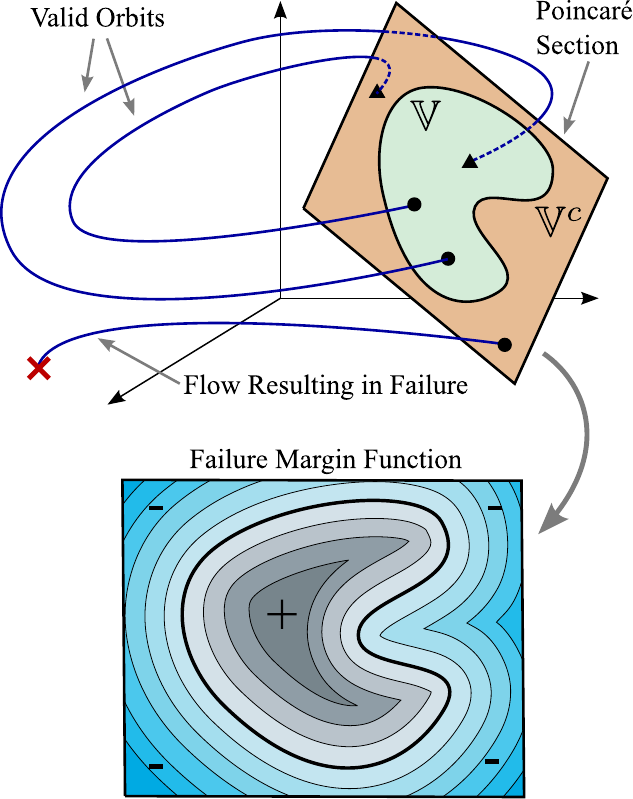}
    \caption{The relationship between the Poincar\`{e} section of a dynamical system and its failure margin function. 
    Legged locomotion is plagued with failure conditions (such as falling over) so we classify points on the Poincar\`{e} section into a set ($\mathbb{V}$) if they produce valid orbits. 
    This set is then used to generate an oriented distance function that we call the failure margin function.\vspace{-6 mm}}
    \label{fig:LeadImage}
\end{figure}


Considerable interest recently has focused on using dynamically rich models for bipedal locomotion planning.
Unfortunately, many more complex models have no closed-form solution and analytical approximations are often unwieldy \cite{Holmes06thedynamics, Haitao2019SLIP3dApprox}.
In \cite{Xiong2020SLIPAmes} the authors planned footsteps with the simple LIP model, mapped those plans to a more dynamic and complex actuated SLIP model then used those plans to control a hardware biped. 
A different approach is to approximate the step-to-step dynamics directly instead of approximating the continuous dynamics.
This has been shown to be effective in a single-step optimal controller \cite{Bhounsule2020Approx} and in a multistep model predictive controller \cite{Zamani2020NMPCNN}.

One area that has not been as well investigated is handling failure conditions of these discrete dynamical models.
Models of legged locomotion have failure modes in which they never return to their cyclic gait, such as when the stance foot slips.
These failure modes result in complex, non-convex failure boundaries in state-action space \cite{Heim2019BeyondBasins}.
Naively bounding the state and action ranges as to not include any failures severely limits the dynamism of potential motions.
Previous work has examined this safety problem by focusing on the continuous-time dynamics, applying tools such as sum-of-squares programming \cite{Jinsun_SafeTemplateAnchor, Nils_ConfidenceSafety} and implicit safety filtering \cite{Gurriet_SafetyCriticalControl}.
Additionally, other work has looked at building safe regions of state-space for learning based control \cite{Wabersich_learningSafety}.

In this work, we propose an optimization-based planning method for analytically intractable models of locomotion that include complex failure conditions.
Our method uses a data-driven, differentiable approximation of the controlled first return map and of a state-action failure margin function, shown in \cref{fig:LeadImage}.
This work is a novel extension of the data-driven (Poincar\`{e}) return maps from \cite{Zamani2020NMPCNN} by pushing into failure-rich domains of motion.
It extends the safe-set construction from \cite{Wabersich_learningSafety} by considering the safety of a state-action pair, not just a state.
Further, in this work we construct an oriented distance function in state-action space to inform the planner about the distance to failure and the direction to the failure boundary.


\section{Modeling of Periodic Gaits}
\label{sec:periodicGaits}
To plan multiple steps ahead for a legged robot, we need a dynamic model of locomotion.
This model can be a simplified model, a full-order simulation of a robot, or even data from the real world robot.
Full order simulations and real-world robots present additional challenges, because of their large state dimensions as discussed in \cref{sec:conclusion}.

Our dynamic model can be described by its configuration and velocity $[ \mathbf{q}, \mathbf{\dot{q}} ] \in TQ$ and control input $\mathbf{u} \in \mathbb{R}^{N_u}$.
These models are hybrid models described by a set of continuous dynamics, hybrid guards and reset maps.
The important feature of these dynamics is that they can create a net displacement through orbits in the other components of state. 

\subsection{Poincar\`{e} Map Discrete Dynamics}
\label{sub:PoincareMapDynamics}
A powerful perspective to analyze these orbits is the Poincar\`{e} section and map.
We define a surface of section $\Gamma$ which is transverse to the flow of our state and which intersects with all gait trajectories of interest,
\begin{equation}
\Gamma \subset TQ.
\end{equation}
This section is one dimension smaller than the full state and can be parameterized by a new, reduced state coordinate $s \in \Gamma$.

We seek to define a function, $\Phi(s,a)$, which represents the controlled first return map.
Our system is an actively controlled system, so instead of a traditional Poincar\`{e} Map, our map is augmented by control actions.
This map will allow us to predict the future state of the system while adjusting control actions (e.g. foot placement).
The dynamical system may have continuously varying actions throughout its orbit, so to rein in the infinite dimensional space of inputs we define a finite set of basis functions for a tractable parameterization.
The parameterized are $a \in \mathbb{A}$.
However, some state-action pairs will fail and never return to the Poincar\`{e} section.
For example, the ground reaction forces could violate the friction cone, cause the foot to slip and the robot to fall.
We can define the set of valid state-action pairs that will successfully return to the section as 
\begin{equation}
\mathbb{V} \subset \Gamma \times  \mathbb{A}.
\end{equation}
Thus, the controlled first return map is 
\begin{equation}
    \Phi: \mathbb{V} \longrightarrow \Gamma.
\end{equation}
which maps a valid state-action pair to the next state on the surface of the section.


\subsection{Failure Margin Function}
\label{sub:FailureMarginFunctionTheory}
We need to easily identify if a candidate state and action will fail, and if it does, determine how to change the state and action to be closer to being a success.
To achieve both of these tasks, we construct a failure margin function which is an oriented distance function \cite{delfour2011shapes} (also called signed distance function), often used in computer graphics \cite{Ma1995LoopDetection}.
To construct this function, we first define the distance function from a point ($x \in \mathbb{R}^N$) to a non-empty set ($A \subset \mathbb{R}^N$) with an associated norm ($\norm{\cdot}$),
\begin{equation}
    d_A(x) = inf\{ \norm{x - y} : y \in A \}.
\end{equation}
The failure margin function is defined as
\begin{equation}
b_{\mathbb{V}^c}(s,a) = d_{\mathbb{V}^c}(s,a) - d_{\mathbb{V}}(s,a)
\label{eq:oriented_distance_function}
\end{equation}
where the complement of $\mathbb{V}$ represents the space of invalid state-action pairs, defined as $\mathbb{V}^c = (\Gamma \times  \mathbb{A}) \setminus \mathbb{V}$.
This function will be positive if the state-action pair is valid and negative if the pair is invalid.
Its magnitude represents the minimum distance to the boundary of success and failure.

\section{Approximation of the Step-to-Step System}
\label{sec:approximationStepToStep}
To facilitate real-time, optimization based planning methods we develop a fast to evaluate, differentiable approximation of the controlled first return map and failure margin function.
The failure margin function allows the optimization to constrain the planned state-action pairs to be in the valid set.

\subsection{Controlled First Return Map Approximation}
\label{sub:approxReturnMap}

For our motion planning optimization we create a data-driven approximation of the true first return map ($\Phi(s,a)$), which we will refer to as $P(s,a)$.
Many structures of approximator could be used, such as polynomial or Gaussian processes but we use feed-forward neural networks.
The neural network is fit to a training data set using standard supervised learning techniques.
The training data set is generated by uniformly sampling state-action pairs and performing the numeric integration of the model dynamics.
This will either fail, in which case we discard the sample, or succeed, in which case we add the initial state, action and final state to the training data set.

\subsection{Failure Margin Function Approximation}
\label{sub:FailureMargin}
The failure margin function approximator is more difficult to create because it cannot be directly sampled.
To generate data for our failure margin function we need a method of calculating the oriented distance function from \cref{eq:oriented_distance_function}.

Our solution is to directly use the definition of the oriented distance function.
We uniformly sample the state-action space and forward simulate.
The result allows us to classify each point as either valid or invalid and build two sets of points.
From these sets we construct two k-d trees, one for valid points and one for invalid points.
K-d trees are space-partitioning data structures which allow for fast nearest neighbor searches in k-dimensional space.
We use these trees to efficiently find the closest valid or invalid point.

Once we have the valid and invalid k-d trees we can sample the failure margin function.
We select a random point in state-action space and forward simulate to the next intersection with the surface of section or to failure.
If our point is a valid state-action pair then $d_\mathbb{V}(s,a) = 0$.
If it is invalid then $ d_{\mathbb{V}^C}(s,a) = 0$.
Now we only need to calculate the distance to the other set which is the canonical use-case for a k-d tree.
This allows us to build a training set to fit an feed-forward neural net approximation ($M(s,a)$) of the failure margin function such that, $M(s,a) \approx b_{\mathbb{V}^c}(s,a)$.

\section{Footstep Planning Optimization Problem}
\label{sec:motionPlanning}

There are many ways to formulate a footstep planning problem depending on the desired behavior.
Here we use a basic formulation which tasks the system with reaching a commanded state $N$ steps in the future.
The problem is provided with $s_0$, the current surface of section state, and $s_{\text{goal}}$, the commanded goal state.
This is an effective formulation for traversal of nominally flat, obstacle free environments with a higher level planner or human commanding target states.
The footstep planning problem is formed as the nonlinear program 
\begin{equation}
  	\begin{aligned}
	    & \underset{x}{\text{minimize}}
	    & & f(x)  \\
	    & \text{subject to}
	    & & h_n(x) = 0, & n \in [0,N-1]\\
	    &&& g_n(x) < 0, & n \in [0,N-1] \\
	    &&& h_{\text{goal}}(x) = 0. 
	    \label{eq:optProblem}
	\end{aligned}
\end{equation}
This optimization problem finds a sequence of $N$ states and actions that are dynamically consistent and reach the final goal state while minimizing the objective $f(x)$.
The decision variable $x = [s_1, s_2, ..., s_N, a_0, a_1, ..., a_{N-1}]$ represents the next $N$ states and actions.
The objective function,
\begin{equation}
    f(x) = \sum_{i=0}^{i = N-1} (s_i - s_{i+1})^T H (s_i - s_{i+1}),
    \label{eq:objective}
\end{equation}
minimizes the squared distance sequential apex states, weighted by $H$
This incentivizes gradual changes in state from step to step.
Inspired by excellent results of a similar planner on Cassie \cite{Apgar2018}, we also test the performance of this motion planning problem without an objective, i.e. $f(x) = 0$.
Removing the objective resulted in significantly faster and more reliable convergence to acceptably smooth motion plans for hardware application.

The approximated forward dynamics are enforced through equality constraints,
\begin{equation}
    h_n(x) = P(s_n, a_n) - s_{n+1}.
\end{equation}
The viability of state-action pairs is enforced through inequality constraints, 
\begin{equation}
    g_n(x) = M(s_n, a_n) + \epsilon,
    \label{eq:margin_constraint}
\end{equation}
which ensures the learned failure margin function is greater that a given threshold value, $\epsilon$.
Finally, the goal constraint enforces that the final state of the optimization matched the commanded goal state ($s_{\text{goal}}$),
\begin{equation}
     h_{\text{goal}}(x) = s_N - s_{\text{goal}} .
\end{equation}
We can implement the Jacobian of the constraints analytically through backpropagation of the approximations, $P(\cdot)$ and $M(\cdot)$.

\section{Illustrative Application}
\label{sec:Application_aSLIP}

\begin{figure*}
    \vspace{4 pt}
    \centering
    \includegraphics[width=0.98\textwidth]{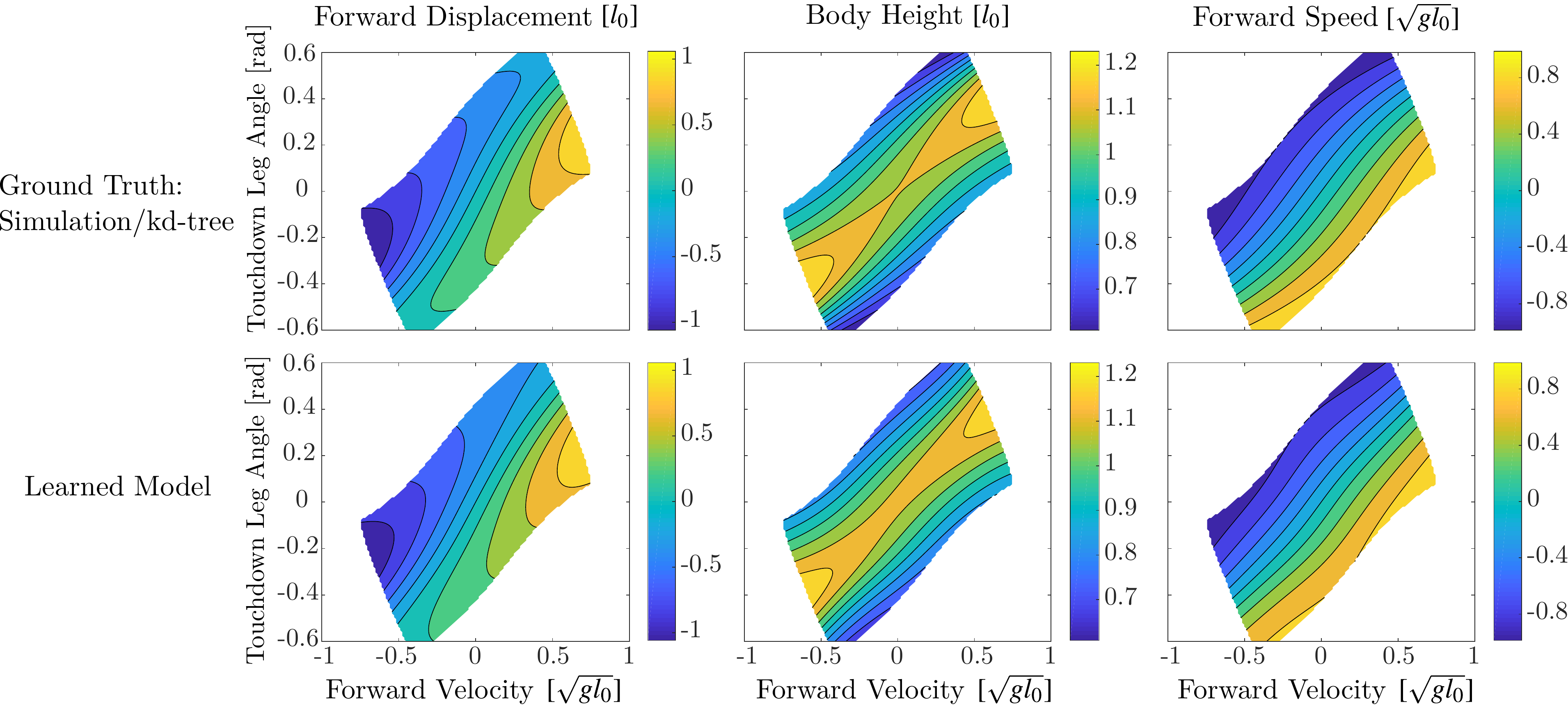}
    \caption{Comparison between ground truth and the approximation of the controlled first return map. For this figure we fixed the initial body height to $1.05 [l_0]$ and the midstance leg extension to $0.05 [l_0]$. The approximator is highly accurate for most of the space, with the largest error near the limits of the input space.\vspace{-5 mm}}
    \label{fig:Model_Slice}
\end{figure*}

\begin{figure}
    \centering
    \includegraphics[width=0.70\columnwidth]{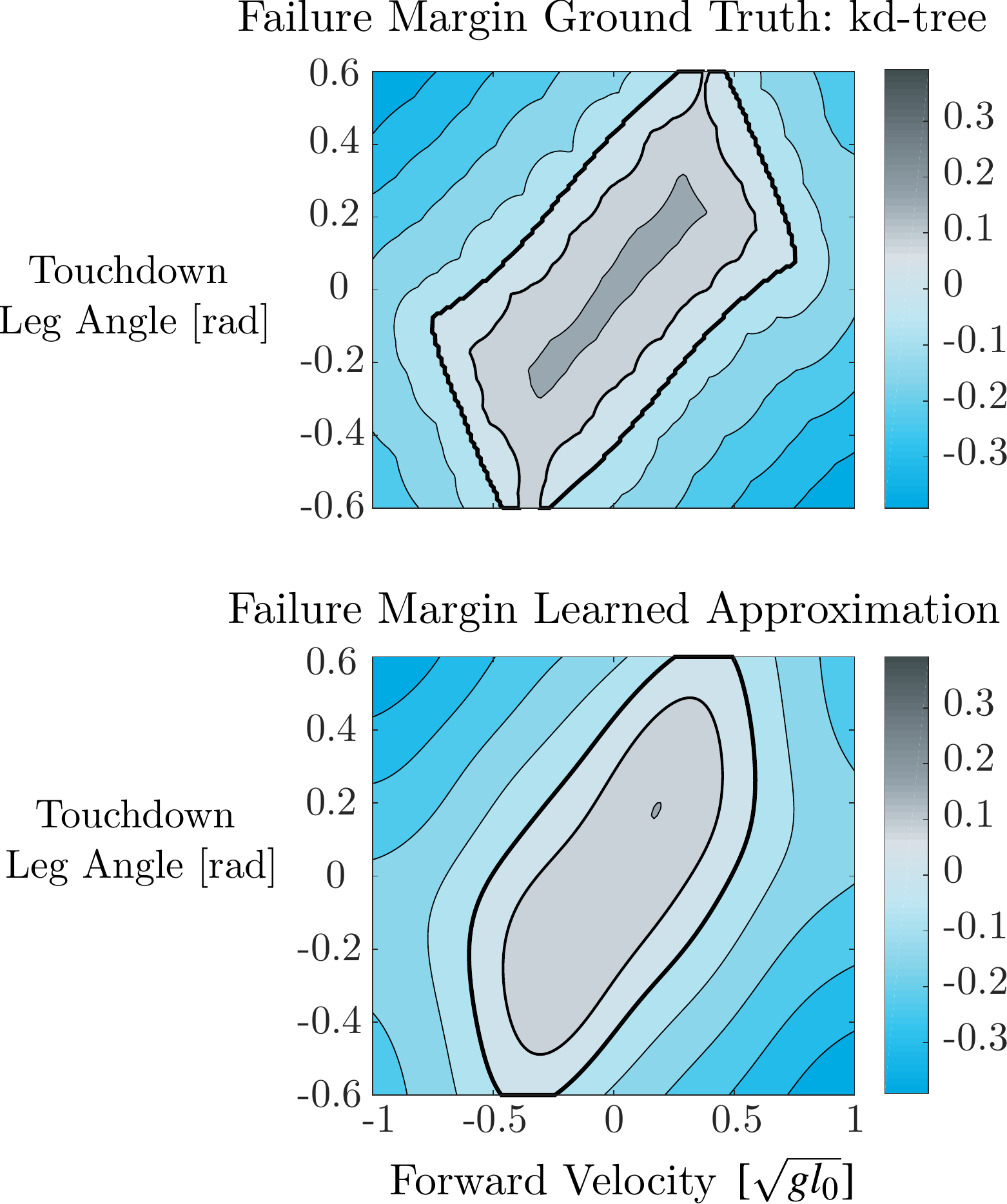}
    \caption{Comparison between ground truth and the learned approximation of the failure margin function.The initial body height is fixed to $1.05 [l_0]$ and the midstance leg extension to $0.05 [l_0]$. The learned approximation captures the general shape of the failure margin function but does not capture all the details of the zero contour (bold contour line).\vspace{-5 mm}}
    \label{fig:Margin_Slice}
\end{figure}
To test the utility and feasibility of this type of planner we choose to test it on one of the simplest systems that demonstrates the problematic features inherent to legged locomotion.
These difficult features include nonlinear dynamics with no closed-form solution, hybrid transitions, failures states and parameterized, low-level control policies.
The simple, illustrative model we use is the nondimensionalized, actuated spring-loaded inverted pendulum (aSLIP) model.
This model consists of a point mass body and a massless leg.
The leg consists of a damped spring in series with an extension actuator which is controlled throughout stance phase and can be instantly repositioned during flight phase.
The nondimensionalization reduced the parameters of the system to only the leg stiffness (20 $[ mgl_0]$), leg damping ratio (0.1) and maximum friction coefficient ($\mu = 0.5$). 
Detailed dynamics of this model and its hybrid transitions are presented in \cite{Green2020PlanningUnexpected}.

The state of the model is described by the position and velocity of the body, $q = [x, y]$, $\dot{q} = [\dot{x}, \dot{y}]$.
This model has failure modes where it falls over or when the foot slips from a friction cone violation.
The inputs are the leg angle at touchdown ($\alpha$) and the motion pattern of the leg extension actuator which is in series with the damped leg spring.
We select a simple actuator pattern inspired by the pneumatic Raibert hopping robots \cite{Raibert_theGoodBook}.
The leg actuation is parameterized by a single value ($\Delta L$) that represents how far the leg actuator will extend at the maximum compression of the leg during stance. 
This highlights a powerful feature of this method, one could design virtually any actuation profile or parameterization that they prefer.
The action could be the true amount of energy to inject or remove, it could be the commanded amount of total energy in the system post extension, or it could be carefully designed to achieve increased robustness such as was shown in \cite{Green2020PlanningUnexpected}. 

\subsection{Step-to-Step Approximation of the aSLIP Model}
\label{sub:aslip_StepToStep}

For the aSLIP model we selected the apex of flight phase as our surface of section.
This is where the vertical velocity is zero during flight phase,
\begin{equation}
\Gamma = \{ [x,y,\dot{x}, \dot{y}] \; : \; \dot{y} = 0 \land \text{Flight Phase}\}.
\end{equation}
This allows us to represent the apex state by the reduced coordinates $s = [x,y,\dot{x}]$.
We can exploit translational invariance to allow us to eliminate the horizontal displacement ($x$) from our apex state for the input.
It's important to know how far each step will translate the robot forward, but the starting location does not change anything about the step or its dynamics.
The final state-action space is
\begin{equation}
[y, \dot{x}, \alpha, \Delta L ] \in \Gamma \times \mathbb{A}.
\end{equation}
We define the limits on the state-action space as
\begin{equation}
\begin{array}{r l l c l l l}
  	0.8     & [l_0]         &\leq &y         &\leq   &1.2   &[l_0] \\
  	-1.0      &[\sqrt{g l_0}] &\leq &\dot{x}   &\leq   &1.0     &[\sqrt{g l_0}] \\
  	-0.6    &[\text{rad}]   &\leq &\alpha    &\leq   &0.6   &[\text{rad}] \\
  	-0.05   &[l_0]          &\leq &\Delta L  &\leq   &0.15  &[l_0].
\end{array}
\label{eq:state-actionLimits}
\end{equation}

The first return map is
\begin{equation}
[\Delta x_{i+1}, y_{i+1}, \dot{x}_{i+1}] = \Phi(y_i, \dot{x}_i, \alpha_i, \Delta L_i),
\end{equation}
where $\Delta x_{i+1}$ is the change in horizontal position from apex $i$ to apex $i+1$.
This is sampled uniformly to create the training data set, form the k-d trees and sample the margin function.
The distance function for state-action space is a weighted 2-norm with weighting matrix of $\text{diag}(6.25, 0.250, 0.309, 2.50)$.
This weighting was chosen to normalize the range of each coordinate in the state-action space.

The neural networks are optimized with the goal of minimizing the weighted 2-norm of the prediction error.
The weighting matrix for prediction error was also chosen to normalize the ranges of the different variables, $\text{diag}(0.250, 6.25, 0.250)$.
The neural networks are implemented using PyTorch and optimized using the ADAM (adaptive momentum estimation) method with the default learning rate of 0.001 \cite{kingma2014adam}.
They were optimized until stationary which took approximately 100,000 iterations which corresponded to 5 to 10 minutes on a computer with an NVIDIA GTX 1080 and an Intel i7-7700k.
We tested several different neural network architectures, varying the width of the hidden layers and activation functions.
We found that there was little difference in the performance between tanh and ReLU activation functions.
Performance improved as the network size increased up to layer widths of 64 neurons.
This led to the final choice to use networks with 2 hidden layers of 64 neurons with ReLU activation functions.


We can examine our return map approximation's performance compared with the contour plots in \cref{fig:Model_Slice}.
These plots show a 2D slice of the 4D state-action space.
The top row shows the ground truth from simulation and the bottom row shows the learned approximator.
We can see that the model performs well for most of the space.
The failure margin function approximator in \cref{fig:Margin_Slice} performs well overall but does not precisely replicate the bold zero margin contour.
The bubble-like arcs in the contour lines are an artifact of individual points in the k-d tree.
These problem would get worse as dimensionality increases, but could both be counteracted with more sophisticated sampling techniques to adaptively add points to the k-d trees near the success/failure boundary.

\begin{figure}
    \centering
    \vspace{4 pt}
    \includegraphics[width=1.0\columnwidth]{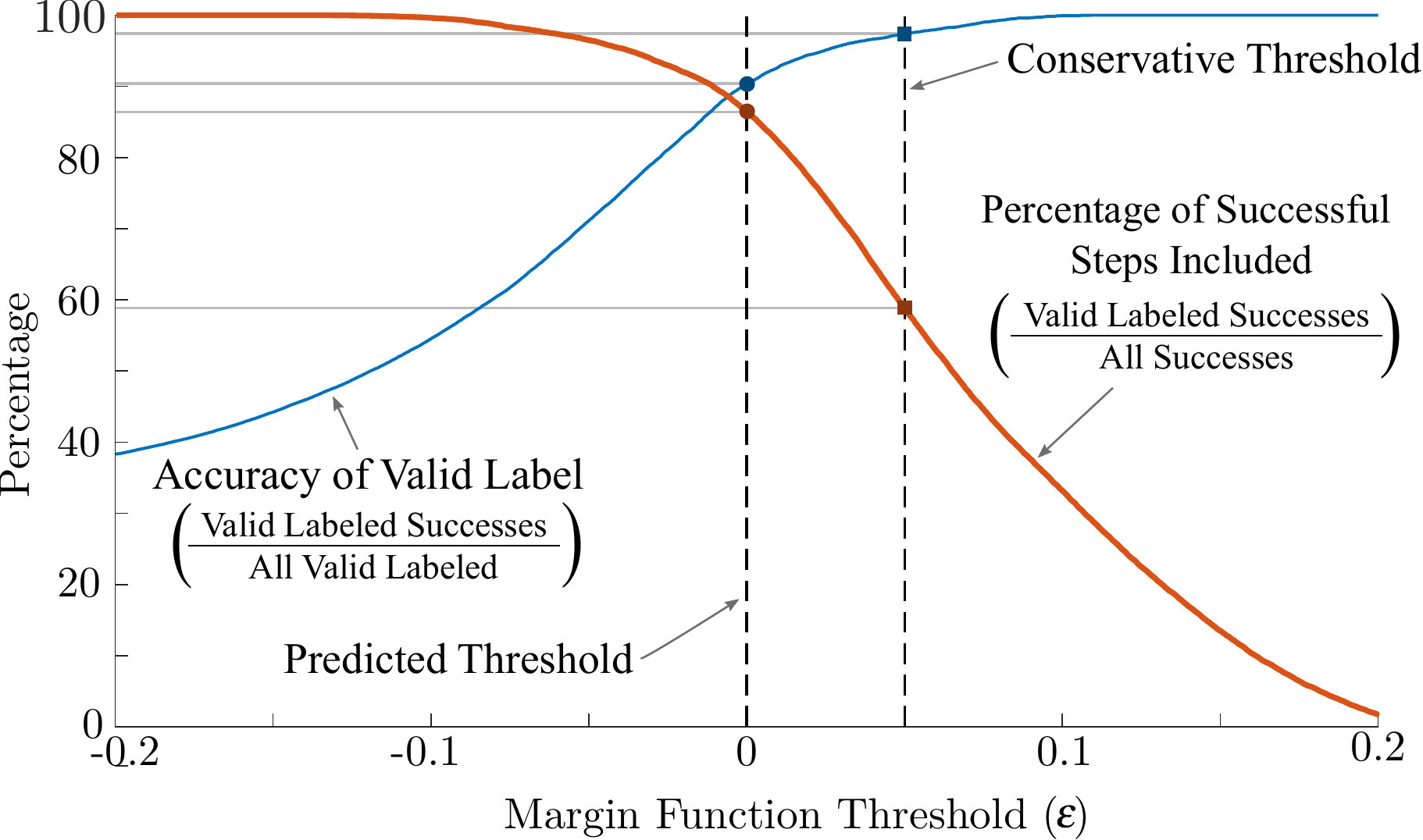}
    \caption{Performance of the failure margin function as we vary the valid point threshold value.
    Increasing the threshold increases the accuracy of the valid label.
    However, increasing the threshold also decreases the percentage of all successful steps that are included.
    We can adjust the threshold to balance the restriction of possible motions with confidence in the accuracy of labeled points.\vspace{-5 mm}}
    \label{fig:Margin_performance}
\end{figure}

The most useful feature of the margin function is accurate classification of state-action points.
We expect a perfect margin function to assign every valid point a positive value and every failure a negative value.
To evaluate the performance, we can look at the classification accuracy as we vary the threshold, $\epsilon$.
This helps us choose a safe threshold for our optimization constraint \cref{eq:margin_constraint}.
However, as we increase the constraint threshold we will exclude more valid points which limits the space of possible behaviors.
\cref{fig:Margin_performance} shows the accuracy of the valid label and the percentage of valid points included as we vary the threshold, $\epsilon$.
From this, we made the engineering decision to use a threshold of $0.05$.
This will result in 97\% of valid labeled points being valid and 59\% of possible valid points being included.

%
%
%
%

\subsection{Footstep Optimization and Failure Margin Utility}
\label{sub:aslip_opt}
To test the utility of the approximations we implemented the multistep locomotion optimization problem from \cref{sec:motionPlanning}.
We assigned our system a random initial state and final state in the range from \cref{eq:state-actionLimits}.
This approach did not command the final position of the robot because it was intended to emulate a gait transition rather than a position-keeping task.
This task is quite difficult and often impossible due to the limited friction coefficient ($\mu = 0.5$).
We implemented the nonlinear optimization problem using the cyipopt python wrapper around Ipopt. \cite{wachter2006ipopt}.
The objective function gradient and constraint Jacobians were analytically calculated using PyTorch's automatic differentiation functionality.


\begin{figure}
    \centering
    \vspace{5 pt}
    \includegraphics[width=0.98\columnwidth]{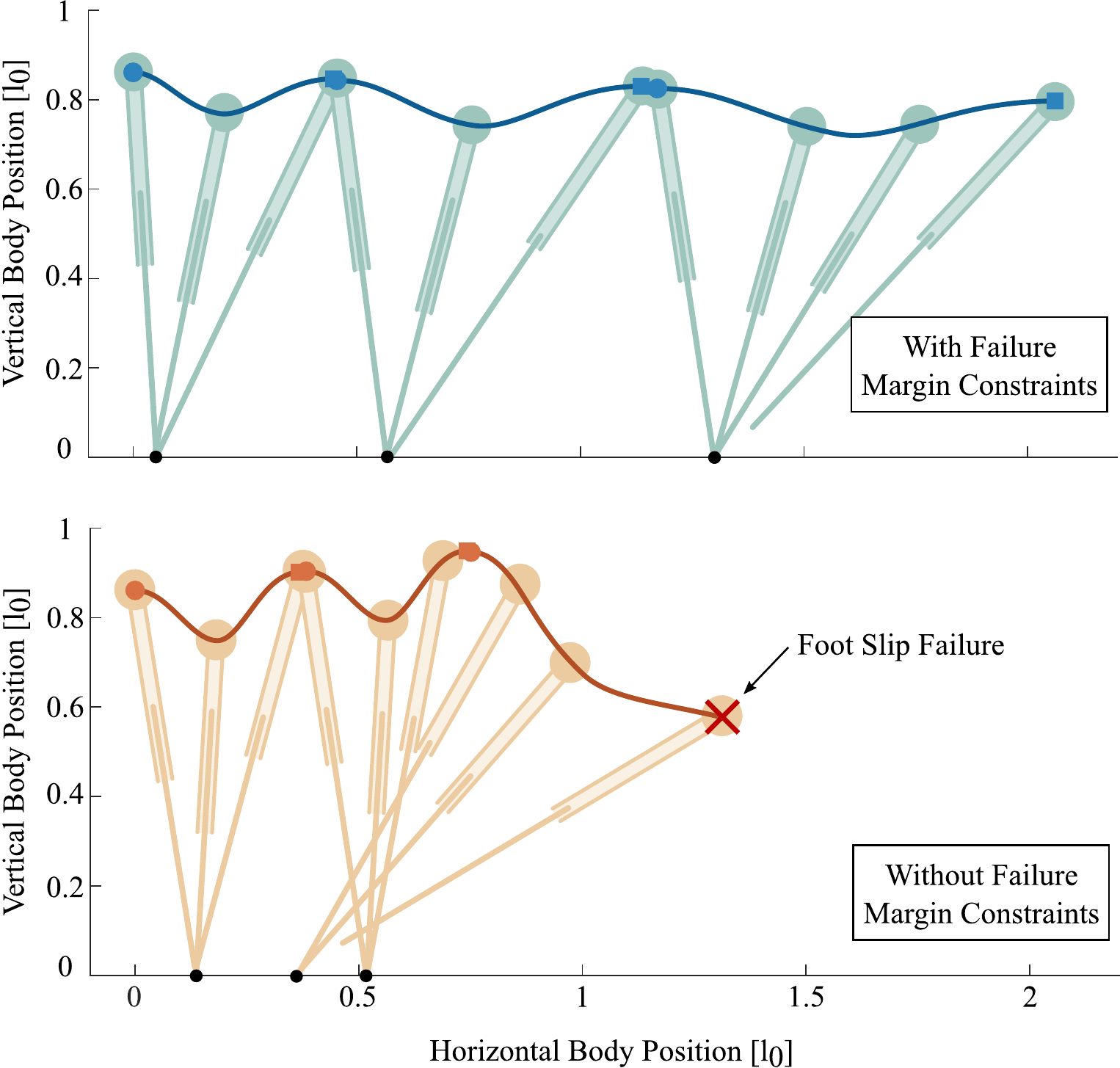}
    \caption{Solutions to an optimization problem given the task to significantly speed up and slightly lower the apex height. The optimization problem with failure margin constraints produce a valid trajectory and the problem without failure margin constraints solves ``successfully" but contains a failure step.\vspace{-6 mm}}
    \label{fig:motion_example}
\end{figure}

Two solutions to a difficult three step task are illustrated in \cref{fig:motion_example}, one with failure margin constraints and one without.
For visualization, we simulated each commanded step and plotted the motion of the body.
The optimization without failure margin constraints thinks it found a valid plan, but we can see that step three results in foot slip as it falls forward.

To examine if the failure margin function improves the reliability of the optimization problem, we ran 1000 motion planning problems with three and four step planning horizons.
The results of these optimizations are summarized in \cref{Tab:OptResults}.
In every situation, the failure margin constraints increased the percentage of problems for which a fully valid solution was found, up to a 22 percentage point increase.
The optimizations with failure margin function constraints declare the problem infeasible more often than those without, but they have a greatly reduced chance of incorrectly returning a solution that contains failure steps (reduced by between 24 and 47 percentage points).
While the time required for optimizations with the failure margin functions is greater, it is only to 1.32 to 2.02 times as long where other safety methods require on average 11 times the time of their unsafe comparisons \cite{Jinsun_SafeTemplateAnchor}.

\begin{table}
\begin{center}
\vspace{4 pt}
\renewcommand{\arraystretch}{1.15}
\setlength{\tabcolsep}{5pt}
\begin{tabular}{|r | r | r | r | r |} 
 \hline
 & \multicolumn{4}{c|}{3 Step Horizon} \\
 \cline{2-5}
 Problem Description & \multicolumn{2}{c|}{State Objective} & \multicolumn{2}{c|}{No Objective} \\
 \cline{2-5}
 & \makecell{Margin} &  \makecell{No Margin} &  \makecell{Margin} & \makecell{No Margin}  \\
 \hline
 Declared Infeasible        & 25.4 \% & 1.1 \%  & 29.0 \%   & 0.5 \%\\ 
  Invalid Solution & 27.6 \% & 56.5 \% & 13.0 \%   & 60.0 \%\\
 Valid Solution         & 47.0 \% & 42.4 \% & 57.6 \%   & 39.5 \% \\ 
 \rule{0pt}{9 pt} Mean Time (sec) & 0.29 & 0.18  & 0.14  & 0.072  \\
 \hline
\end{tabular}
\newline
\vspace*{5 pt}
\newline
\begin{tabular}{|r | r | r | r | r |} 
 \hline
 & \multicolumn{4}{c|}{4 Step Horizon} \\
 \cline{2-5}
 Problem Description & \multicolumn{2}{c|}{State Objective} & \multicolumn{2}{c|}{No Objective} \\
 \cline{2-5}
 & \makecell{Margin} &  \makecell{No Margin} &  \makecell{Margin} & \makecell{No Margin}  \\
 \hline
 Declared Infeasible        & 29.4 \% & 10.5 \%     & 22.6 \%   & 0.4 \%\\ 
  Invalid Solution & 27.8 \% & 51.9 \%     & 11.9 \%   & 56.1 \%\\
 Valid Solution         & 42.8 \% & 37.6 \%     & 65.5 \%   & 43.5 \% \\
 \rule{0pt}{9 pt} Mean Time (sec)        & 1.29 & 0.98  & 0.16  & 0.079  \\
 \hline
\end{tabular}
\end{center}
\caption{Motion planning performance on 1000 random tasks.\vspace{-10 mm}}
\label{Tab:OptResults}
\end{table}


%
%
%

\section{Conclusions}
\label{sec:conclusion}
In this paper we propose a method to plan motions for computationally intractable, failure prone models of locomotion through a novel failure margin function.
This approach was demonstrated on a canonical model of locomotion that is well understood but does not have a closed form solution.
The failure margin function constraints decreased the frequency of invalid motion plans by between 24 and 47 percentage points.
Use of the failure margin function increased the average optimization time by between 1.32 and 2.02 times, which is significantly less of an impact compared to other safe motion planning methods \cite{Jinsun_SafeTemplateAnchor}.
The best performing problems are near practical motion planning with computation times that would correspond to 7 Hz replanning frequencies. 
For human scale locomotion, normal step frequencies is at most 2 Hz so having at least five replanning cycles per step is a realistic possibility.

While we demonstrated this problem on a canonical model of locomotion, there are no inherent barriers to extending this to real, full-dimensional robots.
This planning method holds promise for application to a more physically grounded model such as the data-driven, reduced-order models proposed in \cite{Chen2020OptimalReducedOrder}.
The complex dynamics that such models exhibit can be encapsulated cleanly into the controlled first return map.
An alternative approach is to start with a stabilizing locomotion controller for a full order robot, such as in \cite{siekmann2021sim, tucker2020preference}, and sampling a Poincar\`{e} section of the full robot state and controller commands.
The dimensionality of this return map is too large to plan with, but it could be encoded to a compressed representation.
This could enable planning valid motions and control policy commands in the compressed state-action space. 

\section*{Acknowledgements}
\small{We would like to thank Helei Duan, Jeremy Dao, Connor Yates, Aseem Saxena, Fanzhou Yu and Prof. Alan Fern for their thoughtful discussions and feedback on this work.}






\bibliographystyle{IEEEtran.bst}
\bibliography{references.bib}

\end{document}